\documentclass[10pt,journal,compsoc]{IEEEtran}

\ifCLASSOPTIONcompsoc
  \usepackage[nocompress]{cite}
\else
  \usepackage{cite}
\fi
\usepackage[colorinlistoftodos]{todonotes}

\usepackage{amsmath}
\usepackage{amssymb}
\usepackage{hyperref}

\usepackage{amsmath,amsfonts,bm}

\def\eqref#1{equation~\ref{#1}}

\def\1{\bm{1}}

\def\rc{{\textnormal{c}}}

\def\vh{{\bm{h}}}

\def\mC{{\bm{C}}}
\def\mD{{\bm{D}}}

\def\mI{{\bm{I}}}

\def\mM{{\bm{M}}}

\def\mP{{\bm{P}}}

\def\mU{{\bm{U}}}

\DeclareMathAlphabet{\mathsfit}{\encodingdefault}{\sfdefault}{m}{sl}
\SetMathAlphabet{\mathsfit}{bold}{\encodingdefault}{\sfdefault}{bx}{n}

\def\emC{{C}}

\def\emM{{M}}

\newcommand{\R}{\mathbb{R}}

\DeclareMathOperator*{\argmin}{arg\,min}

\begin{document}

\title{Completely Self-Supervised Crowd Counting\\via Distribution Matching}

\author{Deepak Babu Sam*,
        Abhinav Agarwalla*,
        Jimmy Joseph,
        Vishwanath A. Sindagi,\\
        R. Venkatesh Babu,~\IEEEmembership{Senior~Member,~IEEE},
        Vishal M. Patel,~\IEEEmembership{Senior~Member,~IEEE}
\thanks{Manuscript received September 8, 2020; revised **** **, ****; accepted **** **, ****. This work was supported by the Science and Engineering Research Board (SERB), Department of Science and Technology (DST), Government of India under the project SB/S3/EECE/0127/2015. (D. B. Sam and A. Agarwalla contributed equally to this work.) (Corresponding authors: D. B. Sam; A. Agarwalla.)\protect\\
D. B. Sam, A. Agarwalla, J. Joseph and R. V. Babu are with the Video Analytics Lab, Department of Computational and Data Sciences, Indian Institute of Science, Bangalore, India (e-mail: deepaksam@iisc.ac.in; agarwallaabhinav@gmail.com; jimmyj005@gmail.com; venky@iisc.ac.in). \protect\\
V. A. Sindagi and V. M. Patel are with the Vision \& Image Understanding Lab, Department of Electrical and Computer Engineering, Johns Hopkins University, Baltimore, USA (e-mail: vishwanathsindagi@jhu.edu; vpatel36@jhu.edu).
}
}

\IEEEtitleabstractindextext{
\begin{abstract}
Dense crowd counting is a challenging task that demands millions of head 
annotations for training models. Though existing self-supervised approaches 
could learn good representations, 
they require some labeled data to map these
features to the end task of density estimation. We mitigate this issue 
with the proposed paradigm of \emph{complete self-supervision}, which does 
not need even a single labeled image. The only input required to train, apart 
from a large set of unlabeled crowd images, is the approximate upper limit 
of the crowd count for the given dataset. Our method dwells on the idea 
that natural crowds follow a power 
law distribution, which could be leveraged to yield error signals for 
backpropagation. A density regressor is first pretrained with self-supervision 
and then the distribution of predictions is matched to the prior by optimizing 
Sinkhorn distance between the two. Experiments show that this results in effective 
learning of crowd features and delivers significant counting 
performance. Furthermore, we establish the superiority of our method in less data 
setting as well. The code and models for our approach is available at {\tt\href{https://github.com/val-iisc/css-ccnn}{https://github.com/val-iisc/css-ccnn}}.
\end{abstract}

\begin{IEEEkeywords}
Unsupervised Learning, Crowd Counting, Deep Learning
\end{IEEEkeywords}}

\maketitle
\IEEEdisplaynontitleabstractindextext
\IEEEraisesectionheading{\section{Introduction}
\label{sec:introduction}}
\IEEEPARstart{T}{he} ability to estimate head counts of dense crowds effectively 
and efficiently serves several practical applications. This has 
motivated deeper research in the field and resulted in a plethora 
of crowd density regressors. These CNN based models deliver excellent 
counting performance almost entirely on the support of fully supervised 
training. Such a data hungry paradigm is limiting the further development 
of the field as it is practically infeasible to annotate thousands of 
people in dense crowds for every kind of setting under consideration. 
The fact that current datasets are relatively small and cover only limited 
scenarios, accentuates the necessity of a better training regime. 
Hence, developing methods to leverage the easily available unlabeled data, 
has gained attention in recent times.

The classic way of performing unsupervised learning revolves around 
autoencoders 
(\nolinebreak\hspace{1sp}\cite{hinton2006reducing,vincent2008extracting,kingma2013auto,makhzani2015winner}). 
Autoencoders or its variants are optimized to predict back their 
inputs, usually through a representational bottleneck. By doing so, 
the acquired features are generic enough that they could 
be employed for solving other tasks of interest. 
These methods have graduated to the more recent framework of self-supervision, 
where useful representations are learned by performing some alternate task for 
which pseudo labels can be easily obtained. For example, in self-supervision 
with colorization approach 
(\nolinebreak\hspace{1sp}\cite{zhang2016colorful,larsson2016learning,larsson2017colorization}), 
a model is trained to predict the color image 
given its grayscale version. One can easily generate grayscale inputs from 
RGB images. Similarly, there are lots of tasks for which labels are freely 
available like predicting angle of rotation from an image 
(\nolinebreak\hspace{1sp}\cite{gidaris2018unsupervised,feng2019self}), solving jumbled scenes 
\cite{noroozi2016unsupervised}, inpainting \cite{pathak2016context} etc. 
Though self-supervision is effective in learning useful representations, they 
require a final mapping from the features to the end task of interest. 
This is thought to be essentially unavoidable as some supervisory signal 
is necessary to aid the final task. For this, typically a linear layer or 
a classifier is trained on top of the learned features using supervision 
from labeled data, defeating the true purpose of self-supervision. 
In the case of crowd counting, one requires training with annotated data 
for converting the features to a density map. To reiterate, the current 
unsupervised approaches might capture the majority of its features from 
unlabeled data, but demand supervision at the end should they be made 
useful for any practical applications.

\begin{figure*}[t]
\centering
\includegraphics[width=0.95\textwidth,height=0.23\textheight]{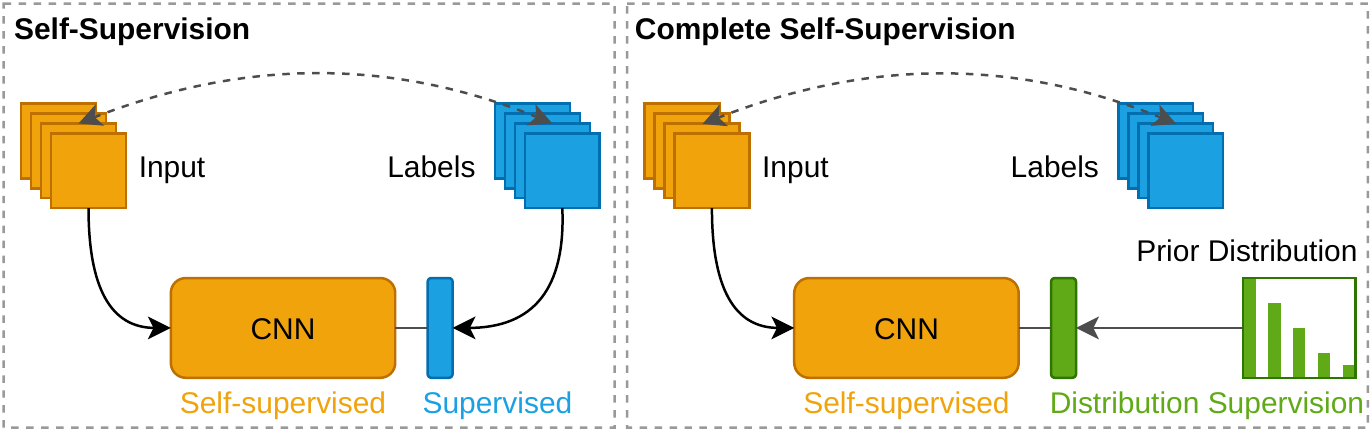}

\caption{Self-Supervision Vs Complete Self-Supervision: Normal self-supervision techniques has a mandatory labeled training stage to map the learned features to the end task of interest (in blue). But the proposed complete self-supervision is devoid of such an instance-wise labeled supervision, instead relies on matching the statistics of the predictions to a prior distribution (in green).}
\label{fig:diff_train_paradigms}
\end{figure*}

Our work emerges precisely from the above limitation of the standard 
self-supervision methods, but narrowed down to the case of crowd density 
estimation. The objective is to eliminate the mandatory final labeled supervision 
needed for mapping the learned self-supervised features 
to a density map output. In other words, we mandate developing a model that 
can be trained without using any labeled data. Such a problem statement is not 
only challenging, but also ill-posed. 
Without providing a supervisory 
signal, the model cannot recognize the task of interest and how to properly guide the 
training stands as the prime issue. We solve this in a novel manner by 
carefully aiding the model to regress crowd density on the back of making some crucial 
assumptions. 
The idea relies on the observation that natural crowds tend to follow 
certain long tailed statistics and could be approximated to an appropriate 
parametric prior distribution (Section \ref{sect:crowd_distribution}). 
If a network trained with a self-supervised task is available (Section 
\ref{sect:stage1_training}), its features 
can be faithfully mapped to crowd density by enforcing the predictions to match 
the prior distribution (Section \ref{sect:stage2_training}). 
The matching is measured in terms of Sinkhorn distance \cite{cuturi2013sinkhorn}, 
which is differentiated to derive error signals for supervision.
This proposed framework is contrasted against the normal self-supervision regime 
in Figure~\ref{fig:diff_train_paradigms}, with the central difference being the 
replacement of the essential labeled training at the end by supervision through 
distribution matching. We show that the proposed approach results in effective 
learning of crowd features and delivers good performance in terms of counting metrics 
(Section \ref{sect:experiments}). 

In summary, our work contributes the following:
\begin{itemize}
\item The first \emph{completely self-supervised} training paradigm which does not 
require instance-wise annotations, but works by matching statistics of the 
distribution of labels.
\item The first \emph{crowd counting model} that can be trained without using a single 
annotated image, but delivers significant regression performance.
\item A detailed analysis on the distribution of persons in dense crowds to reveal the  \emph{power law nature} and enable the use of optimal transport framework. 
\item A novel extension of the proposed approach to \emph{semi-supervised setting} that can effectively exploit unlabeled data and achieve significant gains.
\item An efficient way to improve the Sinkhorn loss by leveraging \emph{edge information} from crowd images.
\end{itemize}

\begin{figure*}[t]
\centering
\includegraphics[width=0.95\textwidth,height=0.22\textheight]{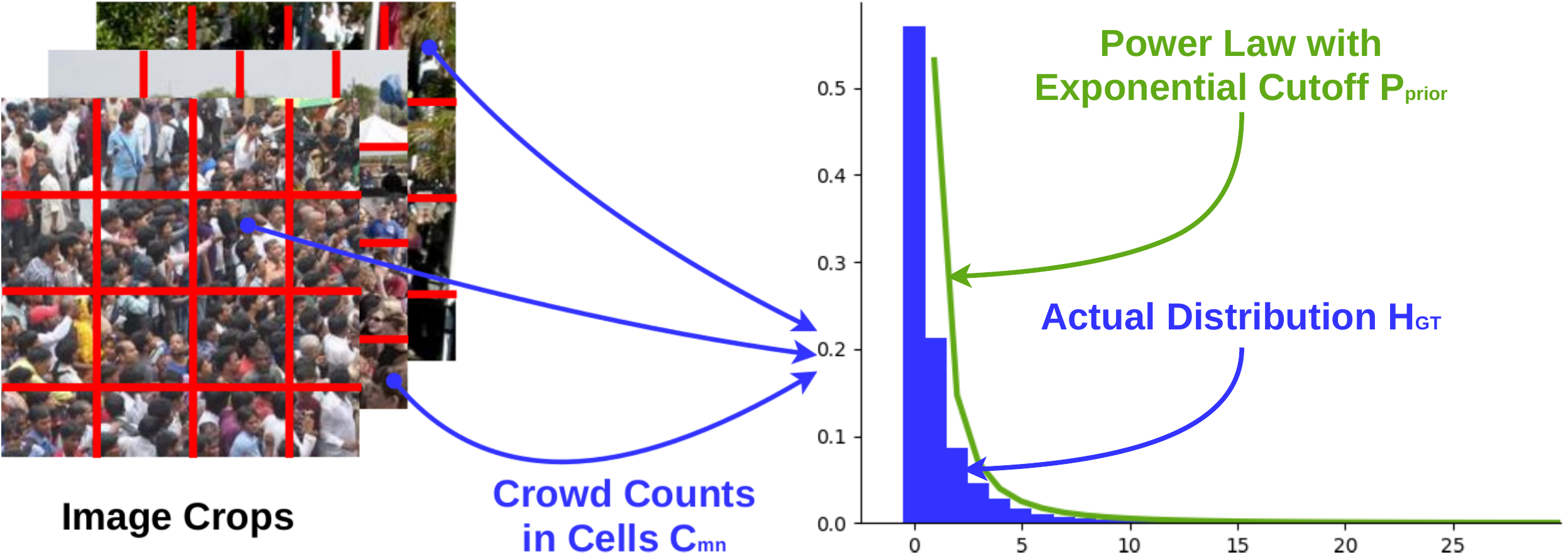}
\caption{Computing the distribution of natural crowds: crops from dense crowd images are framed to a spatial grid of cells and crowd counts of all the cells are aggregated to a histogram (obtained on Shanghaitech Part\_A dataset \cite{zhang2016single}). The distribution is certainly long tailed and could be approximated to a power law.}
\label{fig:crowd_distribution}
\end{figure*}

\section{Related Work}

The paradigm of dense crowd counting via density regression plausibly begins 
with \cite{idrees2013multi}, where hand-crafted features and frequency analysis 
are employed. With the advent of deep learning, many CNN based density regressors 
have emerged. It ranges from the initial simple models 
\cite{zhang2015cross} to multi-network/multi-scale architectures 
designed specifically to address the drastic diversity in crowd images 
(\nolinebreak\hspace{1sp}\cite{onoro2016towards,zhang2016single,sam2017switching,Sam_2018_CVPR,Cao_2018_ECCV}).
Regressors with better, deeper and recurrent based deep models 
(\nolinebreak\hspace{1sp}\cite{Li_2018_CVPR,jiang2019crowd,liu2019recurrent}) are shown to improve 
counting performance. An alternate line of works enhance density regression by providing 
auxiliary information through crowd classification 
(\nolinebreak\hspace{1sp}\cite{sindagi2017cnn,sindagi2017generating}), scene context 
(\nolinebreak\hspace{1sp}\cite{AAAItdfcnn,liu2019context,cheng2019learning}), perspective data 
(\nolinebreak\hspace{1sp}\cite{yan2019perspective,shi2019revisiting}), attention 
(\nolinebreak\hspace{1sp}\cite{zhang2019relational,zhang2019attentional,liu2019adcrowdnet}) and even 
semantic priors \cite{wan2019residual}. Models designed to progressively predict 
density maps and perform refinement is explored in 
(\nolinebreak\hspace{1sp}\cite{Ranjan_2018_ECCV,Idrees_2018_ECCV,sindagi2019pushing}). Works like 
(\nolinebreak\hspace{1sp}\cite{liu2019crowd,sindagi2019multi}) effectively fuse multi-scale information. A better ground truth density map should 
result in better regression and (\nolinebreak\hspace{1sp}\cite{wan2019adaptive,xu2019learn}) leverage such refinement 
opportunities. Some recent approaches try to bring flavours of detection to crowd counting 
(\nolinebreak\hspace{1sp}\cite{Liu_2018_CVPR,lian2019density,liu2019point,sam2020dotdetect,sam2020locate}).
Developing alternate loss functions is also an area of focus with multi-task loss 
formulations like (\nolinebreak\hspace{1sp}\cite{zhao2019leveraging,shi2019counting}) and probablistic training 
regimes as in \cite{ma2019bayesian}. Some counting works employ Negative Correlation Learning 
\cite{Shi_2018_CVPR}, adversarial training \cite{Shen_2018_CVPR} and divide-and-conquer 
approach \cite{xiong2019open}. Interestingly, all these works are fully supervised and 
leverage annotated data to achieve good performance. The issue of annotation has drawn 
attention of a few works in the field and is mitigated via multiple means. 
A count ranking loss on unlabeled images is employed in a multi-task formulation along with 
labeled data by \cite{unlabelled_liu_tpami}. Wang et al. \cite{wang2019learning} train 
using labeled synthetic data and adapt to real crowd scenario. The autoencoder method 
proposed in \cite{almost_unsup} optimizes almost 99\% of the model parameters with 
unlabeled data. However, all of these models require some annotated data 
(either given by humans or obtained through synthetic means) for training, which we aim to eliminate.

Our approach is not only new to crowd counting, but also kindles alternate avenues 
in the area of unsupervised learning as well. Though initial works on the subject 
employ autoencoders or its variants 
(\nolinebreak\hspace{1sp}\cite{hinton2006reducing,vincent2008extracting,kingma2013auto,makhzani2015winner}) 
for learning useful features, the paradigm of self-supervision with pseudo labels 
stands out to be superior in many aspects. Works like 
(\nolinebreak\hspace{1sp}\cite{zhang2016colorful,larsson2016learning,larsson2017colorization}), learn representations through colourising a grayscale image. Apart from these, pseudo 
labels for supervision are computed from motion cues 
(\nolinebreak\hspace{1sp}\cite{agrawal2015learning,jayaraman2015learning,pathak2017learning}), 
temporal information in videos (\nolinebreak\hspace{1sp}\cite{wang2015unsupervised,misra2016shuffle}),
learning to inpaint \cite{pathak2016context}, co-occurrence \cite{isola2015learning},
spatial context (\nolinebreak\hspace{1sp}\cite{noroozi2016unsupervised,doersch2015unsupervised,nathan2018improvements}), 
cross-channel prediction \cite{zhang2017split}, spotting artifacts \cite{jenni2018self}, 
predicting object rotation (\nolinebreak\hspace{1sp}\cite{gidaris2018unsupervised,feng2019self})
etc. The recent work of Zhang et al. \cite{zhang2019aet} introduce the idea of 
auto-encoding transformations rather than data. An extensive and rigorous comparison 
of all major self-supervised methods is available in \cite{kolesnikov2019revisiting}.
All these approaches focus on learning generic features and not the final task. But we 
extend the self-supervision paradigm further directly to the downstream task of interest.

\section{Our Approach\label{sect:our_approach}}

\subsection{Natural Crowds and Density Distribution\label{sect:crowd_distribution}}
As mentioned in Section \ref{sec:introduction}, our objective of training a density regressor 
without using any annotated data is somewhat ill-posed. The main reason being the absence of any 
supervisory signal to guide the model towards the task of interest, which is the density estimation 
of crowd images. But this issue could be circumvented by effectively exploiting certain structure 
or pattern specific to the problem. In the case of crowd images, restricting to only dense 
ones, we deduce an interesting pattern on the density distribution. They seem to spread out 
following a power law. To see this, we sample fixed size crops from lots of dense crowd 
images and divide each crop into a grid of cells as shown in Figure~\ref{fig:crowd_distribution}. 
Then the number people in every cell is computed and accumulated to a histogram. The distribution 
of these cell counts is quite clearly seen to be long tailed, with regions having low counts 
forming the head and high counts joining the tail. The number of cell regions with no people 
has the highest frequency, which then rapidly decays as the crowd density increases. This 
resembles the way natural crowds are arranged with sparse regions occurring more often than 
rarely forming highly dense neighborhoods. Coincidentally, it has been shown that many natural 
phenomena obey a similar power law and is being studied heavily \cite{clauset2009power}.
The dense crowds also, interestingly, appears conforming to this pattern as evident from 
multiple works (\nolinebreak\hspace{1sp}\cite{helbing2007dynamics, moussaid2011simple, karamouzas2015universal, helbing2009power} etc.) on the dynamics of pedestrian gatherings. 

Moving to a more formal description, if $\mD$ represents the density map for the  
input image $\mI$, 
then the crowd count is given by $C=\sum_{xy}\mD_{xy}$ (please refer 
(\nolinebreak\hspace{1sp}\cite{idrees2013multi,zhang2016single,sam2017switching} etc.) regarding creation of 
density maps). $\mD$ is framed into a grid of $M\times N$ (typically set as $M=N=3$) 
cells, with $\emC_{mn}$ 
denoting the crowd count in the cell indexed by $(m,n)$. Now let $H^{GT}$ be the 
histogram computed by collecting the cell counts ($\emC_{mn}$s) from all the images. 
We try to find 
a parametric distribution that approximately follows $H^{GT}$ with special focus to 
the long tailed region. The power law with exponential cut-off seems to be better suited
(see Figure~\ref{fig:crowd_distribution}). Consequently, the crowd counts in cells 
$\emC_{mn}$ could be thought as being generated by the following relation, 
\begin{equation}
\emC_{mn}\thicksim P_{prior}(\rc)\propto c^{\alpha}\exp(-\lambda c),
\end{equation}
where $P_{prior}$ is the substitute power law distribution. There are two parameters to 
$P_{prior}$ with $\alpha$ controlling the shape and $\lambda$ setting the tail length. 

\begin{figure*}[t]
\centering
\includegraphics[width=\textwidth]{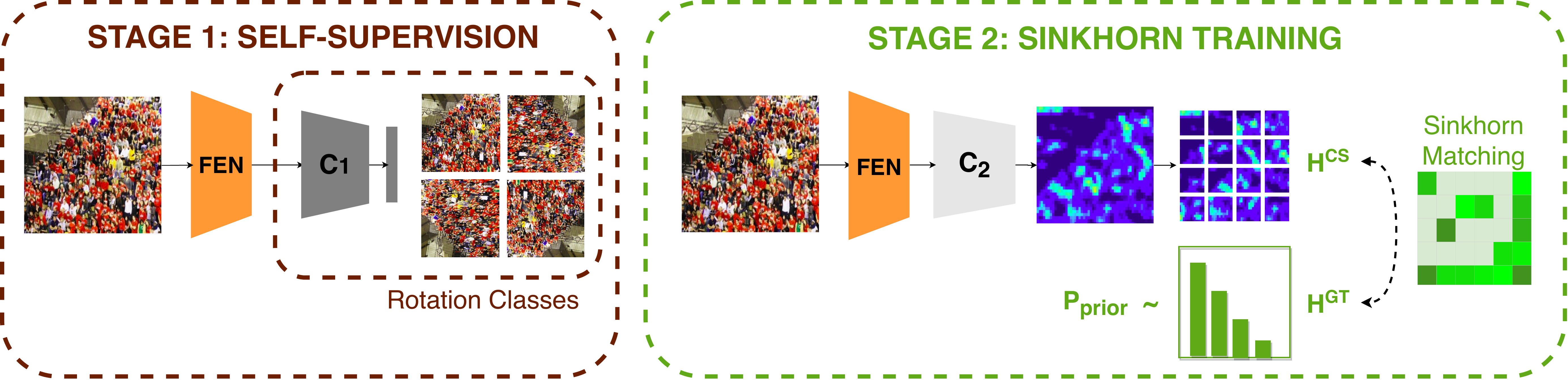}
\caption{The architecture of CSS-CCNN is shown. CSS-CCNN has two stages of training: the first trains the base \emph{feature extraction network} in a self-supervised manner with rotation task and the second stage optimizes the model for matching the statistics of the density predictions to that of the prior distribution using optimal transport.}
\label{fig:cuccnn_architecture}
\end{figure*}

Our approach is to fix a prior distribution so that it can be enforced on the 
model predictions. Studies like (\nolinebreak\hspace{1sp}\cite{helbing2007dynamics, moussaid2011simple}, etc.) simulate crowd behaviour dynamics and estimate the exponent of the power law to be around $2$. 
Empirically, we also find that $\alpha=2$ works in most cases of dense crowds, with 
the only remaining parameter to fix is the $\lambda$. Observe that $\lambda$ affects the 
length of the tail and directly determines the maximum number of people in any given cell. 
If the maximum count $C^{max}$ is specified for the given set of crowd images, then $\lambda$ 
could be fixed such that the cumulative probability density (the value of CDF) of $P_{prior}$ 
at $C^{max}$ is very close to $1$. We assume $1/S$ as the probability of finding a cell 
with count $C^{max}$ out of $S$ images in the given set. Now the CDF value at $C^{max}$ could 
be set to $1-1/S$, simply the probability for getting values less than the maximum. Note 
that $C^{max}$ need not be exact as small variations do not change $P_{prior}$ significantly. 
This makes it practical as the accurate maximum count might not be available in real-world 
scenarios. Since $C^{max}$ is for the cells, the maximum crowd count of the full image  
$C^{fmax}$ is related as $C^{max}=C^{fmax}/(MN S_{crop})$, where $S_{crop}$ denotes the 
average number of crops that make up a full image (and is typically set as 4). 
Thus, for a given a set of highly dense images, 
only one parameter, the $C^{fmax}$ is required to fix an appropriate prior distribution. 

We make a small modification to the prior distribution $P_{prior}$ as its value range starts 
from $1$. $H^{GT}$ has values from zero with large probability mass concentrated near the 
low count region. Roughly 30\% of the mass is seen to be distributed for counts less than or 
around $1$. So, that much probability mass near the head region of $P_{prior}$ is redistributed 
to $[0,1]$ range in a uniform manner. This is 
found to be better for both 
training stability and performance. 

In short, now we have a prior distribution representing how the crowd density is being 
allocated among the given set of images. Suppose there exists a CNN model that can output 
density maps, then one could try to generate error signals for updating the parameters of 
the model by matching the statistics of the predictions with that of the prior. But that 
could be a very weak signal for proper training of the model. It would be helpful if the 
model has a good initialization to start the supervision by distribution matching, which 
is precisely what we do by self-supervision in the next section.

\subsection{Stage 1: Learning Features with Self-Supervision\label{sect:stage1_training}}

We rely on training the model with self-supervision to learn effective and generic 
features that could be useful for the end task of density estimation. That means the 
model has to be trained in stages, with the first stage acquiring patterns frequently 
occurring in the input images. Since only dense crowd images are fed, we hope to 
learn mostly features relevant to crowds. These could be peculiar edges discriminating 
head-shoulder patterns formed by people to fairly high-level semantics pertaining to crowds. 
Note that the model is not signaled to pick up representations explicitly pertinent to density 
estimation, but implicitly culminate in learning crowd patterns as those are the most 
prominent part of the input data distribution. Hence, the features acquired by 
self-supervision could serve as a faithful initialization for the second stage of 
distribution matching.

Regarding self-supervision, there are numerous ways to generate pseudo labels for 
training models. The task of predicting image rotations is a simple, 
but highly effective for learning good representations \cite{kolesnikov2019revisiting}. 
The basic idea is to randomly rotate an image and train the model to predict the angle of 
rotation. By doing so, the network learns to detect characteristic edges or even fairly 
high-level patterns of the objects relevant for determining the orientation. These features 
are observed to be generic enough for diverse downstream tasks \cite{kolesnikov2019revisiting} 
and hence we choose self-supervision through rotation as our method.

Figure \ref{fig:cuccnn_architecture} shows the architecture of our density regressor, 
named the CSS-CCNN (for \emph{Completely Self-Supervised Counting CNN}). It has a base 
\emph{Feature Extraction Network} (FEN), which is composed of three VGG 
\cite{simonyan2014very} style convolutional blocks with max poolings in-between. This 
is followed by two task heads: $C_1$ for the first training stage of self-supervision, and $C_2$ for regressing crowd density at second stage. The first stage branch has two more convolutions and a fully connected layer to finally 
classify the input image to one of the rotation classes. We take $112\times 112$ crops 
from crowd images and randomly rotate the crop by one of the four predefined angles 
(0, 90, 180, 270 degrees).
The model is trained with cross-entropy loss between the predicted and the actual rotation 
labels. The optimization runs till saturation as evaluated on a validation set of images.

Once the training is complete, the FEN has learned useful features for density estimation 
and the rotation classification head is removed. Now the parameters of FEN are frozen and 
is ready to be used in the second stage of training through matching distributions.

\subsection{Stage 2: Sinkhorn Training\label{sect:stage2_training}}

After the self-supervised training stage, FEN is extended to a density regressor by 
adding two convolutional layers as shown in Figure \ref{fig:cuccnn_architecture}. 
We take features from both second and third convolution blocks for effectively mapping to 
crowd density. This aggregates features from slightly different receptive fields and 
is seen to deliver better performance. The layers of FEN are frozen and only a few 
parameters in the freshly added layers are open for training in the second stage of 
distribution matching. This particularly helps to prevent over-fitting as the training 
signal generated could be weak for updating large number of parameters. Now we describe 
the details of the exact matching process.

The core idea is to compute the distribution of crowd density predicted by CSS-CCNN 
and optimize the network to match that closely with the prior $P_{prior}$. For this, a 
suitable distance metric between the two distributions should be defined with 
differentiability as a key necessity. Note that the predicted distribution is in the 
form of an empirical measure (an array of cell count values) and hence it is difficult 
to formulate an easy analytical 
expression for the computing similarity. The classic Earth Mover's Distance (EMD) 
measures the amount of probability mass that needs to be moved if one tries to transform 
between the distributions (also described as the optimal transport cost). But this is not 
a differentiable operation per se and cannot be used directly in our case. Hence, we choose 
the Sinkhorn distance formulation proposed in \cite{cuturi2013sinkhorn}. Sinkhorn distance 
between two empirical measures is proven to be an upper bound for EMD and 
has a differentiable implementation. Moreover, this method performs favorable in terms of efficiency and speed as well. 

Let $\mD^{CS}$ represent the density map output by CSS-CCNN and $\mC^{CS}$ 
hold the cells extracted from the predictions. To make the distribution matching statistically 
significant, a batch of images are evaluated to get the cell counts ($\mC^{CS}_{mn}$s), which 
are then formed into an array $H^{CS}$. We also sample the prior $P_{prior}$ and create 
another empirical measure $H^{GT}$ to act as the ground truth. Now the Sinkhorn loss 
$\mathcal{L}_{sink}$ is computed between $H^{GT}$ and $H^{CS}$. It is basically a 
regularized version of optimal transport (OT) distance for the two sample sets. Designate 
$\vh^{GT}$ and $\vh^{CS}$ as the probability vectors (summing to 1) associated with the 
empirical measures $H^{GT}$ and $H^{CS}$ respectively. Now a transport plan $\mP$ 
could be conceived as the joint likelihood of shifting the probability mass from 
$\vh^{GT}$ to $\vh^{CS}$. Define $\mU$ to be the set of all such valid candidate 
plans as,
\begin{equation}
\mU=\{\mP\in\R_+^{d\times d}\; |\; \mP\1=\vh^{GT}, \mP^T\1=\vh^{CS}\}.
\end{equation}
There is a cost $\mM$ associated with any given transport plan, 
where $\emM_{ij}$ is the squared difference between the counts of $i$th sample of $H^{GT}$ 
and $j$th of $H^{CS}$. Closer the two distribution, lower would be the cost for transport. 
Hence, the Sinkhorn loss $\mathcal{L}_{sink}$ is defined as the cost pertinent to the 
optimal transportation plan with an additional regularization term. Mathematically,
\begin{equation}\label{eqn:sink_loss}
\mathcal{L}_{sink}(H^{GT},H^{CS}) = \argmin_{\mP \in \mU}\left <\mP,\mM\right>_{F} - \frac{1}{\beta} E(\mP),
\end{equation}
where $<>_{F}$ stands for the Frobenius inner product, $E(\mP)$ is the entropy of the joint distribution $\mP$ and $\beta$ is a
regularization constant (see \cite{cuturi2013sinkhorn} for more details). It is evident 
that minimizing $\mathcal{L}_{sink}$ brings the two distributions closer in terms of 
how counts are allotted.

The network parameters are 
updated to optimize $\mathcal{L}_{sink}$, thereby bringing the distribution of predictions 
close to that of the prior. At every iteration of the training, a batch of crowd images are 
sampled from the dataset and empirical measures for the predictions as well as prior are 
constructed to backpropagate the Sinkhorn loss. The value of $\mathcal{L}_{sink}$ on a 
validation set of images is monitored for convergence and the training is stopped if the 
average loss does not improve over a certain number of epochs. Note that we do not use any 
annotated data even for validation. The counting performance is evaluated 
at the end with the model chosen based on the best mean validation Sinkhorn loss.

Thus, our Sinkhorn training procedure does not rely on instance-level supervision, but 
exploits matching the statistics computed from a set of inputs to that of the prior. 
One criticism regarding this method could be that the model need not learn the task 
of crowd density estimation by optimizing the Sinkhorn loss. It could learn any other 
arbitrary task that follows a similar distribution. The counter-argument stems from 
the semantics of the features learned by the base network. Since the initial training 
mostly captures features related to dense crowds (see Section \ref{sect:stage1_training}), 
the Sinkhorn optimization has only limited flexibility in what it can do other than map 
them through a fairly simple function to crowd density. This is especially true as there 
is only a small set of parameters being trained with Sinkhorn. It is highly likely and 
straightforward to map the frequent crowd features to its density values, whose 
distribution is signaled through the prior. Moreover, we show through extensive 
experiments in Section \ref{sect:experiments} and \ref{sect:ablations} 
that CSS-CCNN actually ends up learning crowd 
density estimation. 

\subsection{Improving Sinkhorn Matching\label{sect:aux_info}}

As described already, the Sinkhorn training updates the network parameters 
by backpropagating Sinkhorn loss $\mathcal{L}_{sink}$, 
which brings the distribution of the density predictions closer to that 
of the prior. But computing $\mathcal{L}_{sink}$ relies on estimating 
the optimal transport plan $\mP^{*}$ (the solution to optimization in  
\eqref{eqn:sink_loss}) through the Sinkhorn iterations (see 
\cite{cuturi2013sinkhorn} for more details). 
The quality of estimation of $\mP^{*}$ directly affects the performance of the 
model. Hence, it is quite beneficial to aid the computation of $\mP^{*}$ by 
providing additional information. Any signal that can potentially boost the 
transport assignments is helpful. For example, even simply grouping the 
prediction measures $H^{CS}$ to a coarse sparse-dense categories and then 
restricting assignments within the groups from that of the prior, 
leads to improved performance. This is because the restricted assignments 
make sure that the dense samples from prediction are always mapped to dense 
points in the prior (similarly for sparse ones), reducing costly errors 
of connecting dense ones to sparse and vice versa.
However, one needs to have the density category information to 
supplement the Sinkhorn assignments and that should be obtained in an 
unsupervised fashion as well. 

We observe that the edge details of crowd images could serve as an indicator 
of density. Highly crowded regions seem to have more density of edges, 
while it is low for relatively sparse or non-crowds. But this is a weak signal and can have 
lots of false positives. The higher density of edges could arise from 
non-crowds such as background clutter or other patterns with more edges. 
Interestingly for dense crowds, we find that this weak supervisory signal 
is good enough for grouping regions into potential dense or very sparse categories. 
For any given crowd image, the standard Canny edge detector \cite{canny1986computational} 
is applied to extract the edge map. The map is then blurred and 
down-sampled to look like density maps. These pseudo maps resemble the actual 
ground truth crowd density in many cases, having a relatively higher response 
in dense region than at sparse ones. Note that the absolute values from 
the pseudo maps do not follow the actual crowd density and 
hence cannot be directly used for supervision. However, given a set of 
crowd patches, the relative density values are sufficient to faithfully 
categorize regions into two broad density groups. This is done by first sorting 
pseudo counts of the patches and then dividing the samples at a predetermined 
percentile. Crowd regions with pseudo count values above this threshold are  
considered to be dense while those below goes to the highly sparse or non-crowd. 
By employing a percentile threshold, accurate count values are not required and 
the pseudo counts should only need to be relatively correct across the given set 
of images. Since any random set of crowd patches should follow the prior distribution 
(as per the assumptions and approximations in Section \ref{sect:crowd_distribution}), 
the percentile threshold is fixed on the prior. We fix the threshold to be 30th 
percentile as there are roughly 30\% samples that are non-crowds or with 
very low counts in the range of 1. 

We modify the Sinkhorn training to incorporate the pseudo density information in the 
following manner: first, we compute the pseudo counts $H^{CSP}$ corresponding 
to the prediction samples $H^{CS}$. Using pseudo counts $H^{CSP}$, $H^{CS}$ 
is split to the sparse part $H^{CS}_{0}$ and the dense $H^{CS}_{1}$. 
The prior samples are also grouped with the same threshold to get 
$H^{GT}_{0}$ and $H^{GT}_{1}$.
Now the Sinkhorn loss is separately found for both the categories and added. 
The exact loss being backpropagated is,
\begin{equation}\label{eqn:pseudo_sink}
\mathcal{L}_{sink}^{++}(H^{GT},H^{CS}) = \mathcal{L}_{sink}(H_{0}^{GT},H_{0}^{CS}) + \mathcal{L}_{sink}(H_{1}^{GT},H_{1}^{CS})
\end{equation}
By separating out the assignment of sparse and dense samples, the counting 
performance of the model increases as evident from the experiments 
in Section \ref{sect:experiments}. Note that the 
the Sinkhorn training is complete on its own without the auxiliary density information. 
It is a simple addendum to the method that can improve the performance.

\section{Experiments\label{sect:experiments}}

\subsection{Evaluation Scheme and Baselines}

Any crowd density regressor is evaluated mainly for the standard 
counting metrics. There are two metrics widely being followed 
by the community. The first is the MAE or Mean Absolute Error, 
which directly measures the counting performance. It is 
the absolute difference of the predicted and actual counts 
averaged over the test set or simply  expressed as 
$\textrm{MAE}=(1/S_{test})\sum_{i=1}^{N}|C_{i}-C_{i}^{GT}|$,
where $C_{i}$ is the count predicted by the model for $i$th image 
and $C_{i}^{GT}$ denotes the actual count. Note that $S_{test}$ is the number 
of images in the test set. Coming to the second metric, the 
Mean Squared Error or MSE is defined as 
$\textrm{MSE}=\textrm{SQRT}((1/S_{test})\sum_{i=1}^{N}(C_{i}-C_{i}^{GT})^{2})$,
a measure of the variance of count estimation and it represents the robustness
of the model.

\begin{figure*}[t]
\centering
\includegraphics[width=\textwidth,height=0.59\textheight]{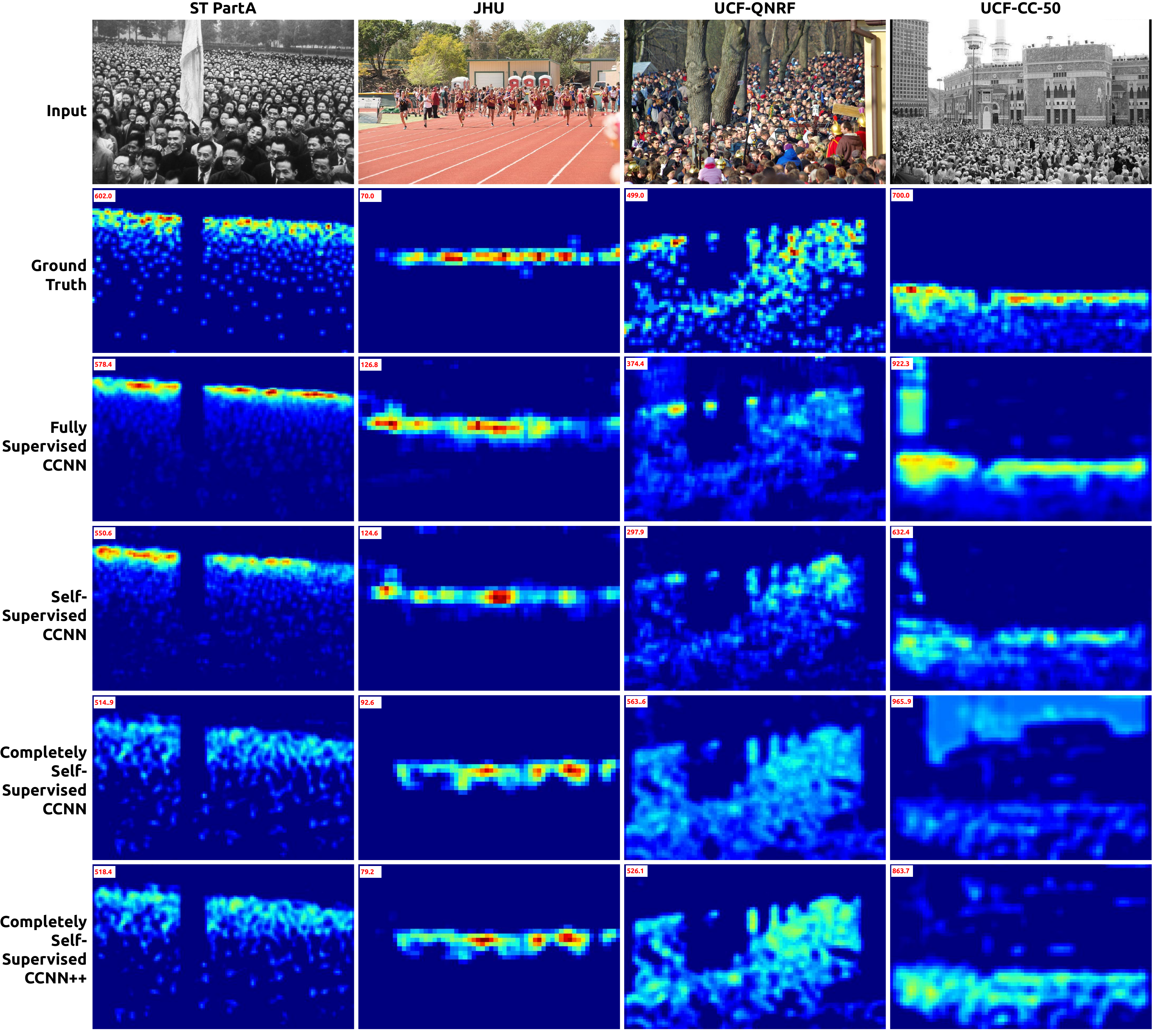}
\caption{Density maps estimated by CSS-CCNN along with that of baseline methods. Despite being trained without a single annotated image, CSS-CCNN is seen to be quite good at discriminating the crowd regions as well as regressing the density values.}
\label{fig:css_predictions}
\end{figure*}

Our completely self-supervised framework is unique in many ways 
that the baseline comparisons should be different from the typical 
supervised methods. It is not fair to compare CSS-CCNN with other 
approaches as they use the full annotated data for training. Hence, we 
take a set of solid baselines for our model to demonstrate its performance. 
The \emph{CCNN Random} experiment refers to the results one would 
get if only \emph{Stage 1} self-supervision is done without the subsequent 
Sinkhorn training. This is the random accuracy for our 
setting and helpful in showing whether the proposed complete 
self-supervision works. Since our approach takes one parameter, 
the maximum count value of the dataset ($C^{fmax}$) as input, 
\emph{CCNN Mean} baseline indicates the counting performance if 
the regressor blindly predicts the given value for all the images. 
We choose mean value as it makes for sense in this setting 
than the maximum (which anyway has worse performance than mean). 
Another important validation for our proposed paradigm is the 
\emph{CCNN $P_{prior}$} experiment, where the model gives out 
a value randomly drawn from the prior distribution as its prediction 
for a given image. The counting performance of this baseline 
tells us with certainty whether the \emph{Stage 2} training does anything 
more than that by chance. Apart from these, the \emph{CCNN 
Fully Supervised} trains the entire regressor with the ground truth 
annotations from scratch. 
Note that we do not initialize CCNN with any pretrained weights 
as is typically done for supervised counting models.
\emph{CCNN Self-Supervised with Labels}, on the 
other hand, runs the \emph{Stage 1} training to learn the FEN parameters 
and is followed by labeled optimization for updating the regressor 
layers. These are not directly comparable to our approach as 
we do not use any annotated data for training, but are shown for completeness.

We evaluate our model on different datasets in the following 
sections. The results for the naive version of the Sinkhorn loss 
$\mathcal{L}_{sink}$ (Section \ref{sect:stage2_training}) is labeled as \emph{CSS-CCNN}, 
whereas \emph{CSS-CCNN++} represents the one with the improved $\mathcal{L}^{++}_{sink}$ (Section \ref{sect:aux_info}).
Note that only the train/validation set images 
are used for optimizing CSS-CCNN and the ground truth 
annotations are never used. The counting metrics are computed 
with the labeled data from the test after the full training.
Unless otherwise stated, we use the same hyper-parameters as 
specified in Section \ref{sect:our_approach}.

\subsection{Shanghaitech Dataset}

\setlength{\tabcolsep}{4pt}
\begin{table}[!b]
\begin{center}
\caption{Performance comparison of CSS-CCNN with other methods on Shanghaitech PartA \cite{zhang2016single}. Our model outperforms all the baselines.}
\label{table:st_parta}
\begin{tabular}{|c|c|c|}
\hline
Method & MAE & MSE\\
\hline
\hline
CCNN Fully Supervised & 118.9 & 196.6\\
\hline
\hline
Sam et al. \cite{almost_unsup} & 154.7  & 229.4\\
\hline
CCNN Self-Supervised with Labels & 121.2 & 197.5\\
\hline
\hline
C-CNN Random & 431.1 & 559.0\\
\hline
C-CNN Mean & 282.8 & 359.9\\
\hline
C-CNN $P_{prior}$ & 272.2 & 372.5\\
\hline
CSS-CCNN (ours) & 207.3 $\pm$ 5.9 & 310.1 $\pm$ 7.7\\
\hline
CSS-CCNN++ (ours) & 195.6 $\pm$ 5.8 & 293.2 $\pm$ 9.3\\
\hline
\end{tabular}
\end{center}
\end{table}
\setlength{\tabcolsep}{1.4pt}

The Shanghaitech Part\_A \cite{zhang2016single} is a popular 
dense crowd counting dataset, containing $482$ images randomly 
crawled from the Internet. It has images with crowd counts as low as 
33 to as high as 3139, with an average of 501. The train set has 
300 images, out of which 10\% is held out for validation. There 
are 182 images testing. The hyper-parameter used for this 
is $C^{fmax}=3000$. We compare the performance of CSS-CCNN with the baselines listed earlier and 
other competing methods in Table \ref{table:st_parta}. The 
metrics for our method is evaluated for three independent runs 
with different initialization and the mean along with variance is 
reported. It is clear that CSS-CCNN outperforms all the baselines 
by a significant margin.  
This shows that the proposed method works better than 
any naive strategies that do not consider the input images. 
With the improved loss, CSS-CCNN++ achieves around 5\% less 
counting error than the naive version due to the more faithful Sinkhorn matching process.
Moreover, the CCNN network with rotation self-supervision 
also beats the model developed in \cite{almost_unsup}. 
It is worthwhile to note that 
the performance of CSS-CCNN is reminiscent of the results of early fully supervised methods with the MAE being better than 
a few of them as well (see Table 2 of \cite{zhang2016single}). 
Figure \ref{fig:css_predictions} visually compares 
density predictions made by CSS-CCNN and other models. The 
predictions of our approach are mostly on crowd regions and 
closely follows the ground truth, emphasizing its ability to discriminate 
crowds well.

\subsection{UCF-QNRF Dataset}

UCF-QNRF dataset \cite{Idrees_2018_ECCV} is a large and diverse collection 
of crowd images with 1.2 million annotations. There are 1535 images with crowd 
count varying from 49 to 12865, resulting in an average of 815 individuals 
per image. 
The dataset offers very high-resolution images with an average resolution of 2013 $\times$ 2902. 
The count hyper-parameter provided to the model is 
$C^{fmax}=12000$. We achieve similar performance trends on UCF-QNRF dataset as well. 
CSS-CCNN outperforms all 
the unsupervised baselines in terms of MAE and MSE as evident from 
Table \ref{table:ucfqnrf}. 
Since the dataset has extreme diversity in terms of crowd density, it is important 
to improve the Sinkhorn matching process and faithfully assign appropriate counts 
across density categories. Owing to the better distribution matching, 
CSS-CCNN++ achieves around 9\% less counting error than CSS-CCNN, despite the 
dataset being quite challenging.

\subsection{UCF-CC-50 Dataset}
UCF\_CC\_50 dataset \cite{idrees2013multi} has just 50 images with 
extreme variation in crowd density ranging from 94 to 4543. The small 
size and diversity together makes this dataset the most challenging. 
We follow the standard 5-fold cross-validation scheme suggested by the
creators of the dataset to report the performance metrics. Since the 
number of images is quite small, the assumption taken for setting the 
prior distribution gets invalid to certain extent. But a slightly 
different parameter to the prior distribution works. We set $\alpha=1$ 
and $C^{fmax}=4000$. Despite being a small and highly 
diverse dataset, CSS-CCNN is able to 
beat all the baselines. The self-supervised MAE is also better than 
the method in \cite{almost_unsup}. These results evidence 
the effectiveness of our method. CSS-CCNN++ improves upon the result significantly in terms of MSE, indicating improved performance on highly dense crowds. 

\setlength{\tabcolsep}{4pt}
\begin{table}[!t]
\begin{center}
\caption{Benchmarking CSS-CCNN on UCF-QNRF dataset \cite{Idrees_2018_ECCV}. Our approach beats the baseline methods in counting performance.}
\label{table:ucfqnrf}
\begin{tabular}{|c|c|c|}
\hline
Method & MAE & MSE\\
\hline
\hline
CCNN Fully Supervised & 159.0 & 248.0\\
\hline
\hline
CCNN Self-Supervised with Labels & 196.8 & 309.3\\
\hline
\hline
CCNN Random & 718.7 & 1036.3\\
\hline
CCNN Mean & 567.1 & 752.8\\
\hline
CCNN $P_{prior}$ & 535.6 & 765.9\\
\hline
CSS-CCNN (ours) & 442.4 $\pm$ 4.2 & 721.6 $\pm$ 13.9\\
\hline
CSS-CCNN++ (ours) & 414.0 $\pm$ 16.3 & 652.1 $\pm$ 15.6\\
\hline
\end{tabular}
\end{center}
\end{table}
\setlength{\tabcolsep}{1.4pt}

\setlength{\tabcolsep}{4pt}
\begin{table}[!t]
\begin{center}
\caption{Performance CSS-CCNN on UCF-CC-50 \cite{idrees2013multi}. Despite being very challenging dataset, CSS-CCNN achieves better MAE than baselines.}
\label{table:ucf50}
\begin{tabular}{|c|c|c|}
\hline
Method & MAE & MSE\\
\hline
\hline
CCNN Fully Supervised & 320.6 & 455.1\\
\hline
\hline
Sam et al. \cite{almost_unsup} & 433.7 & 583.3\\
\hline
CCNN Self-Supervised with Labels & 348.8 & 484.3\\
\hline
\hline
CCNN Random & 1279.3 & 1567.9\\
\hline
CCNN Mean & 771.2 & 898.4\\
\hline
CCNN $P_{prior}$ & 760.0 & 949.9\\
\hline
CSS-CCNN (ours) & 564.9 & 959.4\\
\hline
CSS-CCNN++ (ours) & 557.0 & 737.9\\
\hline
\end{tabular}
\end{center}
\end{table}
\setlength{\tabcolsep}{1.4pt}

\subsection{JHU-CROWD++ Dataset}

JHU-CROWD++ (\nolinebreak\hspace{1sp}\cite{sindagi2019pushing,sindagi2020jhucrowd++}) is a comprehensive dataset with 
1.51 million head annotations spanning 4372 images. The crowd scenes 
are obtained under various scenarios and weather conditions, making it 
one of the challenging dataset in terms of diversity. Furthermore, JHU-CROWD++
has a richer set of annotations at head level as well as image level. 
The maximum count is fixed to $C^{fmax}=8000$. 
The performance trends are quite similar to other datasets, with our 
approach delivering better MAE than the baselines as evident from 
Table \ref{table:jhu}. This indicates 
the generalization ability of CSS-CCNN across different types of 
crowd datasets.

\setlength{\tabcolsep}{4pt}
\begin{table}[!t]
\begin{center}
\caption{Evaluation of CSS-CCNN on JHU-CROWD++ \cite{sindagi2019pushing} dataset.}
\label{table:jhu}
\begin{tabular}{|c|c|c|}
\hline
Method & MAE & MSE\\
\hline
\hline
CCNN Fully Supervised & 128.8 & 415.9\\
\hline
CCNN Self-Supervised with Labels & 147.5 & 436.2\\
\hline
\hline
CCNN Random & 320.3 & 793.5\\
\hline
CCNN Mean & 316.3 & 732.3\\
\hline
CCNN $P_{prior}$ & 302.3 & 707.621\\
\hline
CSS-CCNN (ours) & 243.6 $\pm$  9.1 & 672.4 $\pm$ 17.1\\
\hline
CSS-CCNN++ (ours) & 197.9 $\pm$ 2.2 & 611.9 $\pm$ 12.0\\
\hline
\end{tabular}
\end{center}
\end{table}
\setlength{\tabcolsep}{1.4pt}

\subsection{Cross Data Performance and Generalization}

In this section, we evaluate our proposed model in a cross dataset 
setting. CSS-CCNN is trained in a completely self-supervised manner 
on one of the dataset, but tested on other datasets. Table 
\ref{table:cross_dataset} reports the MAEs for the cross dataset 
evaluation. It is evident that the features learned from 
one dataset are generic enough to achieve reasonable 
scores on the other datasets, increasing the practical utility of
CSS-CCNN. The difference in performance mainly stems from the changes 
in the distribution of crowd density across the datasets. 
This domain shift is drastic in the case of UCF-CC-50 
\cite{idrees2013multi}, especially since the dataset has 
only a few images.

\setlength{\tabcolsep}{4pt}
\begin{table}[!t]
\begin{center}
\caption{Cross dataset performance of our model; the reported entries are the MAEs obtained for CSS-CCNN and CSS-CCNN++ respectively.}
\label{table:cross_dataset}
\begin{tabular}{|c|c|c|c|}
\hline
Train $\downarrow$ / Test $\rightarrow$ & ST\_PartA & UCF-QNRF & JHU-CROWD++\\
\hline
\hline
ST\_PartA & 207.3, 195.6 & 468.1, 472.4 & 254.0, 251.3\\
\hline
UCF-QNRF & 251.2, 235.7 & 442.4, 414.0 & 236.5, 220.6\\
\hline
JHU-CROWD++ & 290.2, 266.3 & 446.2, 417.4 & 243.6, 197.9\\
\hline
\end{tabular}
\end{center}
\end{table}
\setlength{\tabcolsep}{1.4pt}

\subsection{CSS-CCNN in True Practical Setting}
The complete self-supervised setting is motivated for scenarios 
where no labeled images are available for training. But till now 
we have been using images from crowd datasets with the annotations being intentionally ignored. 
Now consider crawling lots of crowd images from the Internet and 
employing these unlabeled data for training CSS-CCNN. For this, we use 
textual tags related to dense crowds and similarity matching with 
dataset images to collect approximately 5000 dense crowd images.
No manual pruning of undesirable images with motion blur, perspective distortion 
or other artifacts is done. 
CSS-CCNN is trained on these images with the same hyper-parameters as that of  
Shanghaitech Part\_A and the performance metrics are 
computed on the datasets with annotations. From Table 
\ref{table:practical_setting}, it is evident that our model is 
able to achieve almost similar or better MAE on the standard crowd 
datasets, despite not using images from those datasets for training. 
This further demonstrates the generalization ability of CSS-CCNN to learn 
from less curated data, emphasizing the practical utility it could facilitate.

\setlength{\tabcolsep}{4pt}
\begin{table}[!t]
\begin{center}
\caption{Evaluating CSS-CCNN in a true practical setting: the model is trained on images crawled from the web, but evaluated on crowd datasets. The counting performance appears similar to that of training on the dataset.}
\label{table:practical_setting}
\begin{tabular}{|c|c|c|c|c|}
\hline
Train on & \multicolumn{2}{c|}{CSS-CCNN} & \multicolumn{2}{c|}{CSS-CCNN++} \\
\cline{2-5}
web images & MAE & MSE & MAE & MSE \\
\hline
\hline
Test on ST\_PartA & 208.8 & 309.5 & 184.2 & 268.8\\
\hline
Test on UCF-QNRF & 450.7 & 755.9 & 422.1 & 699.9\\
\hline
Test on JHU-CROWD++ & 241.2 & 706.8 & 231.0 & 660.1\\
\hline
\end{tabular}
\end{center}
\end{table}
\setlength{\tabcolsep}{1.4pt}

\subsection{Performance with Limited Data}

\begin{figure*}[!t]
\centering
\includegraphics[width=\textwidth,height=0.26\textheight]{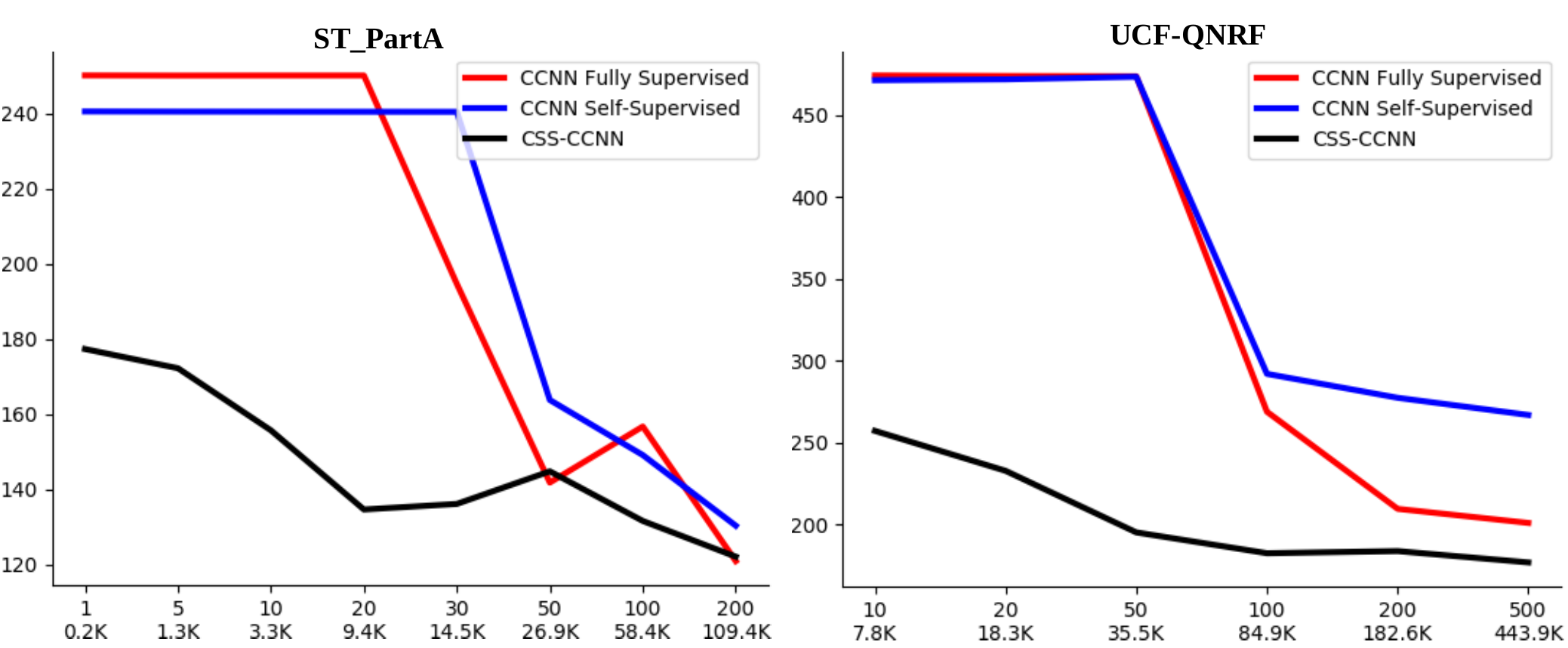}
\caption{Comparing our completely self-supervised method to fully supervised and self-supervised approaches under a limited amount of labeled training data. The x-axis denotes the number of training images along with the count (in thousands) of head annotations available for training, while the y-axis represents the MAE thus obtained. At low data scenarios, CSS-CCNN has significantly superior performance than others.}
\label{fig:mae_vs_data_graph}
\end{figure*}

Here we explore the proposed algorithm along with 
fully supervised and self-supervised approaches 
when few annotated images are available for training. The analysis 
is performed by varying the number of labeled samples 
and the resultant counting metrics are presented in Figure \ref{fig:mae_vs_data_graph}. 
For training CSS-CCNN with data, we utilise the available annotated data to compute the optimal Sinkhorn assignments
$\mP^{*}$ and then optimize the $\mathcal{L}_{sink}$ loss. This way both the 
labeled as well as unlabeled data can be leveraged for training by alternating 
respective batches (in a 5:1 ratio). 
It is clear that, 
at very low data, scenarios CSS-CCNN beats the supervised as well as 
self-supervised baselines by a significant margin. The Sinkhorn training 
shows 13\% boost in MAE (for Shanghaitech Part\_A) by using just one labeled sample as opposed to no samples. This indicates that CSS-CCNN can perform well in extremely low data regimes. It takes about 20K head annotations for the supervised model to perform as well as CSS-CCNN. Also, CSS-CCNN has significantly less number of parameters to learn using the labeled samples as compared to a fully supervised network. These results suggests that our complete 
self-supervision is the right paradigm to employ for crowd counting when the amount of available annotated 
data is less.

\section{Ablations and Analysis\label{sect:ablations}}

\subsection{Ablations on Architectural Choices}

In Table \ref{table:arch_ablation}, we validate our architectural 
choices taken in designing CSS-CCNN. The first set of experiments 
ablates the \emph{Stage 1} self-supervised training. We perform 
Sinkhorn training on a randomly initialized FEN (labeled as 
\emph{Without Stage 1}) and receive a worse MAE. 
In the chosen setting of self-supervision with rotation, 
the input image is randomly rotated by one of the four predefined 
angles for creating pseudo labels. Now we analyse 
the effect of the number of rotation classes on the final counting 
metrics. As evident from the table, 
four angles stands to be the best in agreement with previous 
research on the same \cite{kolesnikov2019revisiting}. Self-supervision 
via colorization is another popular strategy for learning useful 
representations. The model is trained to predict the a-b color 
space values for a given gray-scale image (the L channel). The end 
performance is observed to be inferior in comparison with that of the 
rotation task.
Another option is to load FEN with ImageNet trained weights (as this is 
a typical way of transfer learning) and then employ \emph{Stage 2}. The 
result (\emph{With ImageNet weights}) is worse than that 
of CSS-CCNN, suggesting that the self-supervised training is crucial 
to learn crowd features necessary for density estimation. 
Furthermore, the base 
\emph{feature extraction network} (FEN) (see Figure 3 in main paper) 
is changed to ResNet blocks and CSS-CCNN is trained as well as 
evaluated (\emph{with ResNet based FEN}). Simple VGG style architecture 
appears to be better for density regression.
\setlength{\tabcolsep}{4pt}
\begin{table}[!b]
\begin{center}
\caption{Validating different architectural design choices made for CSS-CCNN evaluated on the Shanghaitech Part\_A \cite{zhang2016single} (computed on single run).}
\label{table:arch_ablation}
\begin{tabular}{|c|c|c|}
\hline
Method & MAE & MSE\\
\hline
\hline
Without \emph{Stage 1} & 257.5 & 397.7\\
\hline
Rotation with 2 class & 233.5 & 344.1\\
\hline
Rotation with 8 class & 232.2 & 341.5\\
\hline
Colorization & 242.5 & 363.0\\
\hline
With ImageNet weights & 257.8 & 370.8 \\
\hline
ResNet based FEN & 244.8 & 332.4 \\
\hline
\hline
Uniform Prior & 261.8 & 406.0\\
\hline
Pareto Prior & 248.3 & 386.2\\
\hline
Lognormal Prior & 239.5 & 345.8\\
\hline
\hline
Without skip connection & 226.8 & 329.1\\
\hline
Cell Size $2\times 2$ & 243.9 & 374.6\\
\hline
Cell Size $4\times 4$ & 251.6 & 389.4\\
\hline
With GT distribution & 202.7 & 300.3\\
\hline
\hline
Percentile threshold 10 & 191.5 & 288.7\\
\hline
Percentile threshold 50 & 189.1 & 286.8\\
\hline
\hline
CSS-CCNN & 197.3 & 295.9\\
\hline
CSS-CCNN++ & 187.7 & 280.21\\
\hline
\end{tabular}
\end{center}
\end{table}
\setlength{\tabcolsep}{1.4pt}
\begin{figure*}[!t]
\centering
\includegraphics[width=\textwidth,height=0.26\textheight]{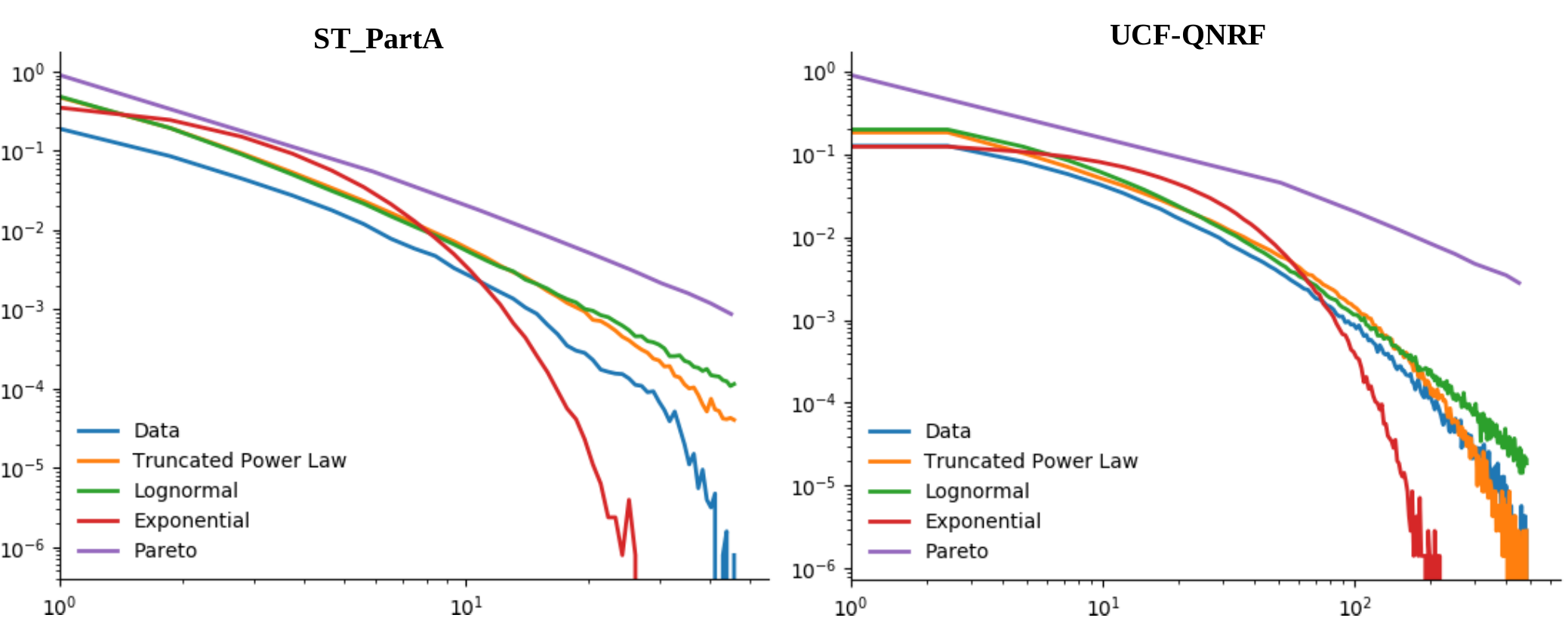}
\caption{Double logarithmic representation of maximum likelihood fit for the crowd counts from Shanghaitech Part\_A \cite{zhang2016single} and UCF-QNRF \cite{Idrees_2018_ECCV}.} 
\label{fig:css_prior_fitness}
\end{figure*}
We also 
run experiments with different types of prior distributions and 
see that the power law with exponential cutoff works better, 
justifying our design choice. The \emph{without skip connection} 
experiment trains CSS-CCNN devoid of the features from the second convolutional block 
in FEN being directly fed to $C_{2}$ (see Figure \ref{fig:cuccnn_architecture} and Section \ref{sect:stage2_training}).
As expected, the feature aggregation from multiple layers improves the 
counting performance. The cell sizes used for computing count 
histograms (see Section \ref{sect:crowd_distribution}) are varied (labeled 
\emph{Cell Size}) to understand the effect on MAE. The metrics 
seem to be better with our default setting of $8\times 8$. 
CSS-CCNN employs a prior parametric distribution to facilitate the 
unlabeled training. We investigate 
the case where the prior is directly given in the form of an empirical 
measure derived from the ground truth annotations. For the Sinkhorn training, 
this \emph{GT distribution} is sampled to get $H^{GT}$ (see Section 
\ref{sect:stage2_training}) 
instead of $P_{prior}$. The resultant MAE is very similar to the 
standard CSS-CCNN setting, indicating that our chosen prior 
approximates the ground truth distribution well. 
Lastly, we ablate the \emph{percentile threshold} used to extract of pseudo 
density category for the CSS-CCNN++ model (Section \ref{sect:aux_info}) 
and find that the default setting helps in better density differentiation.

\subsection{Analysis of the Prior Distribution}

The proposed Sinkhorn training requires a prior distribution of crowd counts 
to be defined and the choice of an appropriate prior is essential for the 
best model performance as seen from Table \ref{table:arch_ablation}. 
Here we analyze the crowd data more carefully to see why the truncated 
power law is the right choice of prior. For this, the counts from 
crowd images are extracted as described in Section \ref{sect:crowd_distribution} and 
a maximum likelihood fit over various parametric distributions is performed. 
The double logarithmic visualization of the probability distribution 
of both the data and the priors are available in Figure \ref{fig:css_prior_fitness}. 
Note that the data curve is almost a straight line in the logarithmic 
plot, a clear marker for power law characteristic. Both 
truncated power law and lognormal tightly follow the distribution. But 
on close inspection of the tail regions, we find truncated power law 
to best represent the prior. This further validates our choice of the 
prior distribution.

\subsection{Sensitivity Analysis for the Crowd Parameter}

As described in Section \ref{sect:crowd_distribution}, CSS-CCNN requires 
the maximum crowd count ($C^{fmax}$) for the given set of images as an 
input. This is necessary to fix the prior distribution parameter $\lambda$. 
One might not have the exact max value for the crowds in a true 
practical setting; an approximate estimate is a more reasonable 
assumption. Hence, we vary $C^{fmax}$ around the actual value and 
train CSS-CCNN on Shanghaitech PartA \cite{zhang2016single} and UCF-QNRF\cite{Idrees_2018_ECCV}. The performance metrics  
in Table \ref{table:cmax_sensitivity} show that changing $C^{fmax}$ 
to certain extent does not alter the performance significantly. The MAE 
remained roughly within the same range, even though 
the max parameter is being changed in the order of 500. Note that the results 
are computed with single runs. These findings indicate that the our approach 
is insensitive to the exact crowd hyper-parameter value, increasing its 
practical utility. We also check the sensitivity of our approach on the 
power law exponent $\alpha$. Varying $\alpha$ around 2 results in similar performances,  
in agreement with the findings of existing works and our design choice 
(see Section \ref{sect:crowd_distribution}).

\setlength{\tabcolsep}{4pt}
\begin{table}[!t]
\begin{center}
\caption{Sensitivity analysis for the hyper-parameters on CSS-CCNN. Our model is robust to fairly large change in the max count parameter.}
\label{table:cmax_sensitivity}
\begin{tabular}{|c|c|c|c|c|c|}
\hline
\multicolumn{3}{|c|}{ST\_PartA} & \multicolumn{3}{c|}{UCF-QNRF}\\
\cline{1-6}
Param & MAE & MSE & Param & MAE & MSE\\
\hline
\hline
$C^{fmax}=2000$ & 204.2 & 316.4 & $C^{fmax}=10000$ & 443.9 & 749.7\\ 
\hline
$C^{fmax}=2500$ & 197.9 & 304.6 & $C^{fmax}=11000$ & 446.9 & 757.5\\
\hline
$C^{fmax}=3000$ & 197.3 & 295.9 & $C^{fmax}=12000$ & 437.0 & 722.3\\
\hline
$C^{fmax}=3500$ & 191.9 & 288.5 & $C^{fmax}=14000$ & 446.1 & 697.5\\
\hline
$\alpha=1.9$ & 202.9 & 303.3 & $\alpha=1.9$ & 438.3 & 700.6\\
\hline
$\alpha=2.0$ & 197.3 & 295.9 & $\alpha=2.0$ & 437.0 & 722.3\\
\hline
$\alpha=2.1$ & 200.7 & 305.6 & $\alpha=2.1$ & 446.4 & 756.3\\
\hline
\end{tabular}
\end{center}
\end{table}
\setlength{\tabcolsep}{1.4pt}

\subsection{Analysis of Features}

\begin{figure*}[!t]
\centering
\includegraphics[width=\textwidth,height=0.22\textheight]{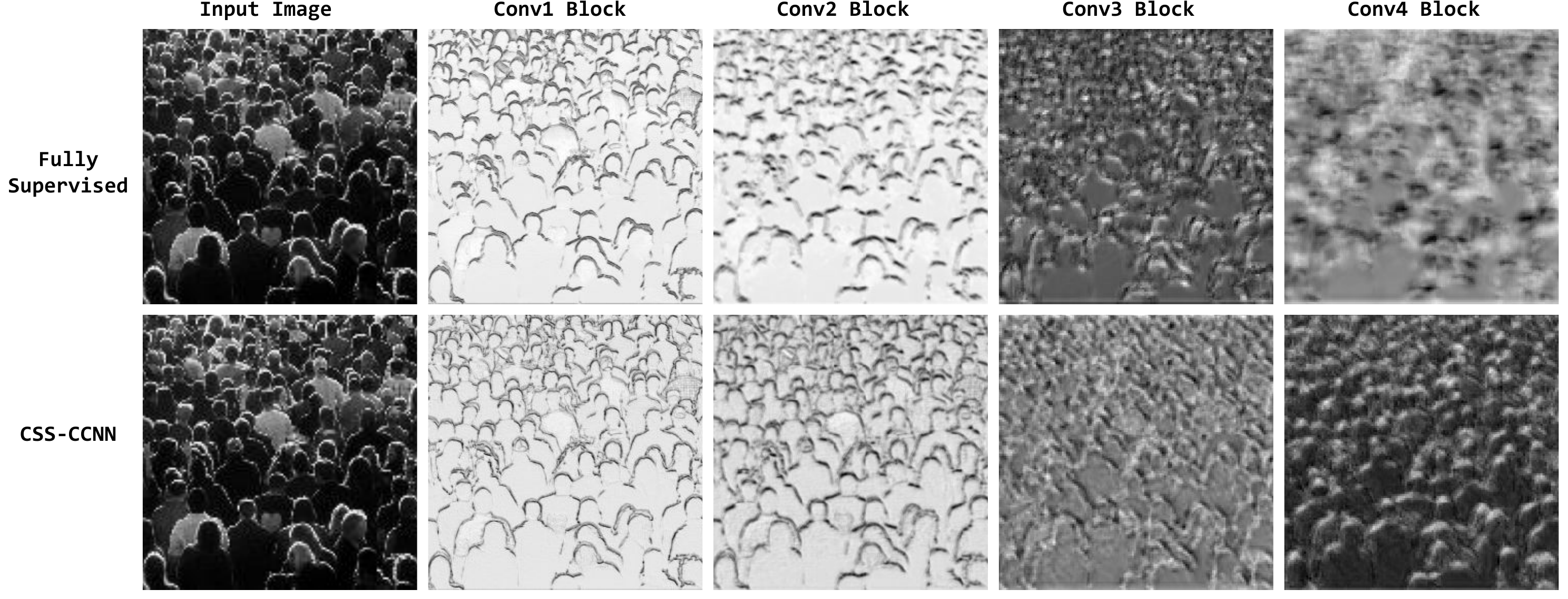}
\caption{Visualization of mean features extracted from different convolutional blocks of CSS-CCNN and the supervised baseline.}
\label{fig:css_features_mean}
\end{figure*}

To further understand the exact learning process of CSS-CCNN, 
the acquired features can be compared against that of a supervised 
model. Figure \ref{fig:css_features_mean} displays the mean feature 
map for the outputs at various convolutional blocks of CSS-CCNN along 
with that of the supervised baseline (see Section \ref{fig:cuccnn_architecture})
evaluated on a given crowd image. Note that Conv4 stands for the regressor 
block that is trained with the Sinkhorn loss in the case of CSS-CCNN. It 
is clear that the self-supervised features closely follow the supervised 
representations, especially at the initial blocks in extracting low-level 
crowd details. Towards the end blocks, features are seen to diverge, with 
fully supervised Conv4 outputs appearing like density maps. But notice that 
the corresponding completely self-supervised outputs have higher 
activations on heads of people, which is relevant for the end task of 
density estimation. This clearly shows that CSS-CCNN indeed learns to 
extract crowd features and detect heads, rather than falling in 
a degenerate case of matching the density distribution without actually 
counting persons.

\section{Conclusions and Future Work}

We show for the first time that a density regressor can be 
fully trained from scratch without using a single annotated image. 
This new paradigm of complete self-supervision relies on optimizing 
the model by matching the statistics of the distribution of 
predictions to that of a predefined prior. 
Though the counting performance of the model stands 
better than other baselines, there is a performance gap compared 
to fully supervised methods. Addressing this issue could be the prime 
focus of future works. For now, our work can  be considered as a proof of 
concept that models could be trained directly for solving the downstream task of 
interest, without providing any instance-level annotated data.

\ifCLASSOPTIONcaptionsoff
  \newpage
\fi

\bibliographystyle{IEEEtran}
\bibliography{references.bib}

\end{document}